\documentclass[10pt,twocolumn]{article}

\usepackage{cvpr}
\usepackage{times}
\usepackage{epsfig}
\usepackage{graphicx}
\usepackage{amsmath}
\usepackage{amssymb}
\usepackage{subfigure}
\usepackage{xcolor}

\usepackage[pagebackref=true,breaklinks=true,colorlinks,bookmarks=false]{hyperref}

\cvprfinalcopy 

\def\cvprPaperID{9} 

\ifcvprfinal\fi
\begin{document}
\title{Feudal Steering: Hierarchical Learning for Steering Angle Prediction}

\author{Faith Johnson\\
Rutgers University\\
{\tt\small faith.johnson@rutgers.edu}
\and
Kristin Dana\\
Rutgers University\\
{\tt\small kristin.dana@rutgers.edu}
}

\maketitle


\begin{abstract}

We consider the challenge of automated steering angle prediction for self driving cars using egocentric road images. In this work, we explore the use of feudal networks, used in hierarchical reinforcement learning (HRL), to devise a vehicle agent to predict steering angles from first person, dash-cam images of the Udacity driving dataset. Our method, Feudal Steering, is inspired by recent work in HRL consisting of a manager network and a worker network that operate on different temporal scales and have different goals. The manager works at a temporal scale that is relatively coarse compared to the worker and has a higher level, task-oriented goal space. Using feudal learning to divide the task into manager and worker sub-networks provides more accurate and robust prediction. Temporal abstraction in driving allows more complex primitives than the steering angle at a single time instance. Composite actions comprise a subroutine or  skill that can be re-used throughout the driving sequence. The associated subroutine id is the manager network's goal, so that the manager seeks to succeed at the high level task (e.g. a sharp right turn, a slight right turn, moving straight in traffic, or moving straight unencumbered by traffic). The steering angle at a particular time instance is the worker network output which is regulated by the manager's high level task. We demonstrate state-of-the art steering angle prediction results on the Udacity dataset. 

\end{abstract}


\section{Introduction}
\thispagestyle{empty}
Reinforcement Learning (RL) has made major strides over the past decade, from learning to play Atari games \cite{mnih2013playing} to mastering chess and Go \cite{silver2017mastering}. However, RL tends to be unable to generalize policies enough to apply them to new environments and still struggles to solve problems with sparse reward signals. In response to this brittleness, Hierarchical Reinforcement Learning (HRL) is growing in popularity. In HRL, a manager network operates at a lower temporal resolution and produces goal vectors that it passes to the worker network. The worker network uses these goal vectors to guide its learning of a policy over micro-actions, also called primitive actions, in the environment at a higher temporal resolution than the manager network \cite{dayan1993feudal}. The temporal abstraction created through this relationship helps the networks to learn and execute macro-actions or tasks, also called {\it subroutines}, in the environment while lessening the negative effects of sparse rewards on network training. 

\begin{figure}
    \centering
    \includegraphics[width=1.02\linewidth]{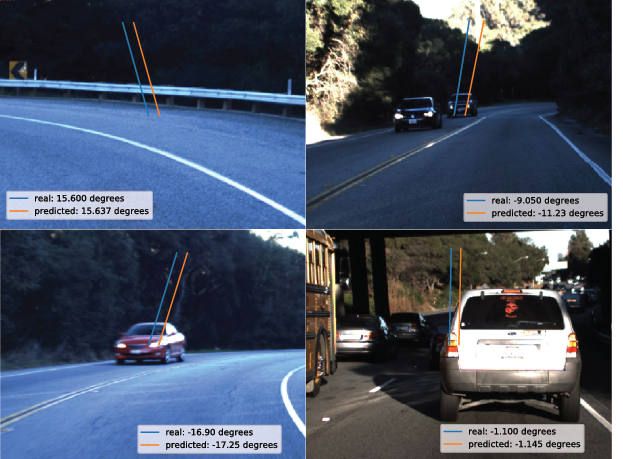}
    \caption{Four frames from the Udacity dataset are shown with their corresponding ground truth (blue) and predicted (orange) steering angles using our Feudal Steering network. The orientation of the lines corresponds to the egocentric steering angle.  Our model predicts steering angles within 2.67 degrees of the ground truth angle. }
    \label{fig:teaser}
\end{figure}

Autonomous driving is an application that struggles with this issue of sparse reward signals. 
However, most HRL work emphasizes video game and other simulated domains instead of autonomous driving applications. At all times, human drivers are paying attention to two levels of their environment. 
The first level goal is on a finer grain: don't hit obstacles in the immediate vicinity of the vehicle. The second level goal is on a coarser grain: plan actions a few steps ahead to maintain the proper course efficiently.
It is even possible to conceive of higher levels of abstraction comprised of path planning and other more complicated driving tasks. 

Autonomous vehicles need to have tight constraints on hardware and software in order to be effective in real world applications  \cite{lin2018architectural}. Current successful HRL networks are large and take a long time to train \cite{song2019diversity,vezhnevets2017feudal}, making them unsuitable to implement in autonomous vehicles despite the theoretical benefits. Additionally, many HRL methods that do focus on the driving domain require handcrafted subroutines and do not focus on primitive navigation directly, choosing to find policies over macro-actions instead. Handcrafting subroutines limits environment exploration and requires a high level of domain specific knowledge in order to yield good model performance. 




We propose a vehicle agent to predict steering angles using feudal networks. Feudal networks are typically applied in hierarchical reinforcement learning. However, in this work, we train these networks with ground-truth data from the Udacity dataset \cite{udacity}, instead of with rewards, allowing us to retain the advantageous hierarchical structure of HRL without using reinforcement learning. We present two methods. The first method predicts steering angles with subroutines (driving tasks) obtained from the t-SNE embedding of the driving data. We also use t-SNE to refine and structure the subroutine embedding space discovered by the manager in order to visualize the driving data subroutines and observe their semantic meaning. The second method allows the manager to discover the existing subroutines in the data instead of handcrafting them.
Our results show that feudal networks with learned subroutines provide improved training stability and prediction performance.


\section{Related Work}
\begin{figure*}
    \centering
    \includegraphics[width=1.02\linewidth]{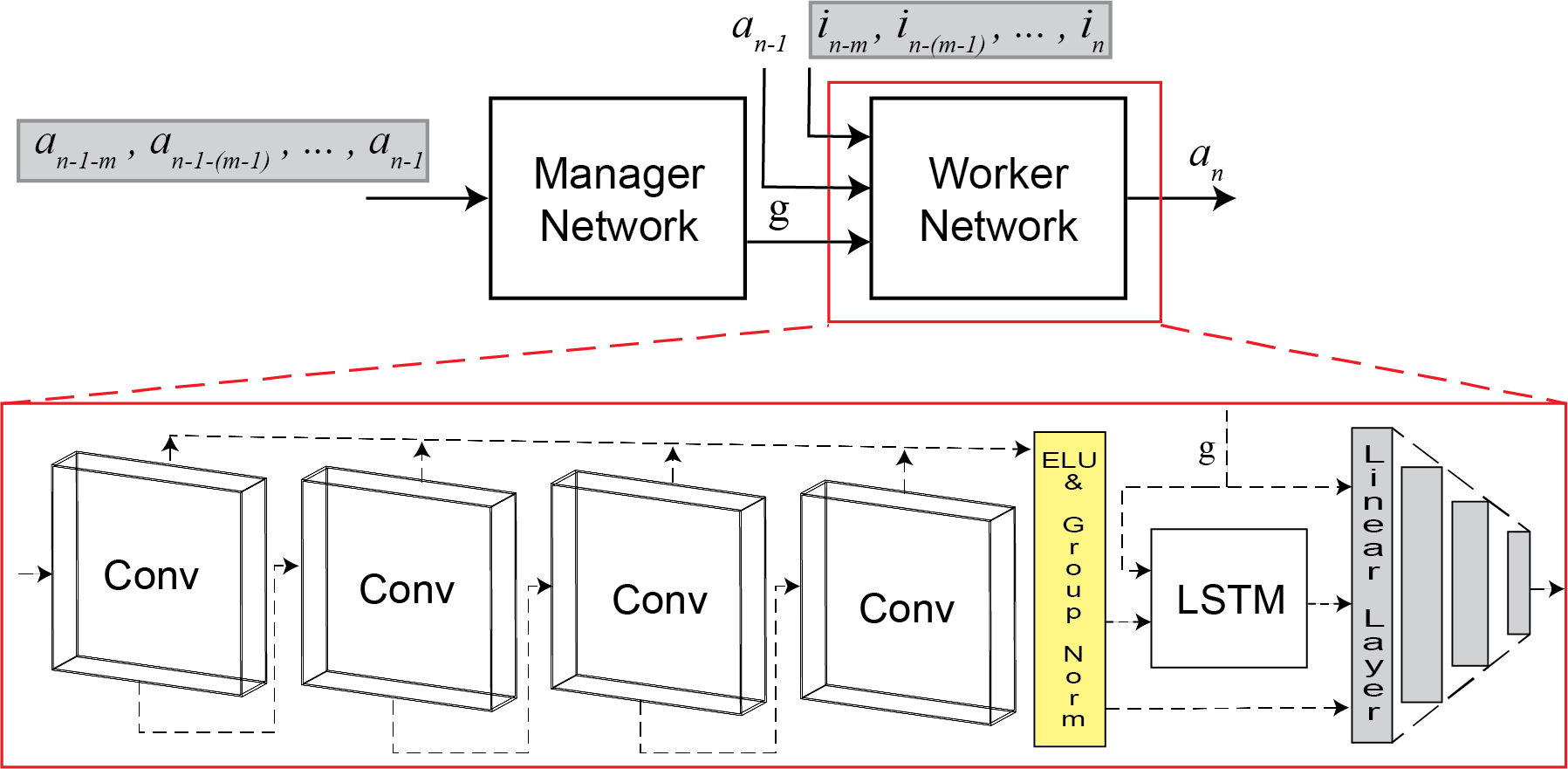}
    \caption{Feudal Steering Network. The overall network  is comprised of a manager network and a worker network. The worker network (expanded in the red box) acts as the  steering angle prediction network. The input to the manager network is a sequence of the previous $m$ predicted steering angles, $[a_{n-1-m}, a_{n-1-(m-1)}, ..., a_{n-1}]$. The input to the worker network is a sequence of $m$ frames, $[i_{n-m}, i_{n-(m-1)}, ..., i_n]$, a goal, $g$, obtained from the manager network, and the previous steering angle, $a_{n-1}$. The yellow box represents the ELU (exponential linear unit) and group normalization step in the pipeline. }
    \label{fig:fullNetwork}
\end{figure*}

\subsection{Temporal Abstraction}
In hierarchical reinforcement learning, the manager network operates at a lower temporal resolution than the worker network 
and communicates with the worker network through a goal vector 
that encapsulates a temporally extended action (called a subroutine, skill, option, or macro-action). 
The worker executes atomic actions in the environment based on this goal vector and its own state information. This process of manager/worker communication through temporal abstraction helps to break down a problem into more 
tractable pieces as outlined by the options framework  \cite{sutton1999between}.

To explain the concept of temporal abstraction further, consider the case of an agent attempting to leave a room through a door. When a human plans this action, they don't compose a low level  sequence of movements such as {\it straight, straight, left, straight, right}. In other words, humans do not consciously think of each atomic action required to exit the room. Instead, they think in terms of temporal abstraction: {\it Find the door. Approach it. Pass through it.} Each of these actions encapsulates multiple atomic actions that need to be executed in a specific order for the agent to complete the higher level task. 

\subsection{Hierarchical Reinforcement Learning}
One difficulty with reinforcement learning is delayed rewards and sparse credit assignment. This problem is especially prevalent 
with RL in autonomous vehicles, as 
an agent may only receive
a reward when it completes a larger sub-task. Hierarchical reinforcement learning is used to increase model performance through temporal abstraction and intrinsic rewards \cite{kulkarni2016hierarchical}, but has limited implementations in the autonomous driving domain as prior work opts for simulated environments. Feudal networks \cite{vezhnevets2017feudal} learns to play the Atari game, Montezuma's Revenge.
Their hierarchical network has a manager that learns a latent space for its goals, 
which take on a directional meaning 
and allow the manager to be updated regardless of the worker's actions in the environment. However, this method requires a lot of data and time to train, which is not necessarily available or possible in the driving domain.

Because of the complexity of the driving domain, there is a trend of manually defining subroutines for HRL networks \cite{chen2018deep,duan2020hierarchical, liang2018cirl}. 
Our method diverges from this practice by allowing the manager network to learn its own subroutines. 
There are other frameworks 
\cite{song2019diversity,tessler2017deep,bacon2017option} that also attempt to learn subroutines implicitly from the data. Kumar et al. \cite{kumar2019learning} propose a method to learn subroutines through imitation learning and propose using HRL to refine them. 
Another approach explores the nature of the subroutines themselves by focusing on learning the states of the subgoals instead of learning the policy between these states \cite{rafati2019learning}. 
This hierarchical approach is taken a step further by \cite{haarnoja2018latent,nachum2018data} which use the states in the latent space of the lower layer as the action space for the layer above it.  



    

\subsection{Steering Angle Prediction}
Most of the work in steering angle prediction uses some form of alternative representation of the driving scene beyond RGB images, from attention maps \cite{kim2017interpretable,he2018aggregated} to segmentation and optical flow \cite{hou2019learning,maqueda2018event, khan2019latent}. While these representations contain valuable information, 
we aim for a method that predicts steering angles
using only raw visual input, as humans do. Additionally, in the case of segmentation and optical flow, these alternative scene representations add latency to the prediction pipeline which is undesirable for real world applications. CNN-based methods such as \cite{liu2019end} use features directly from the RGB image input and use multiple fully connected layers to predict steering angle, speed, and acceleration, thereby allowing them to create a fully functional, end to end, autonomous vehicle model. 

In order to create an autonomous driving system that is robust to real world driving scenarios, it is desirable that real world data is used to train and test the networks as in \cite{bojarski2016end}, that deploys their implementation in a vehicle.
The most comparable steering angle prediction methods to Feudal Steering are \cite{chi2017deep, xu2017end}, which use a sequence of RGB images 
to predict steering angles using recurrent units. 
However,
our approach demonstrates the effectiveness of feudal learning for steering angle prediction by estimating subroutines (macro-action states) across the driving data. 

\section{Methods}

\subsection{Steering Angle Prediction Network}
Our approach to predicting steering angles is inspired by  \cite{komanda} from the Udacity steering angle challenge.  During training, this network inputs images to a CNN to extract the relevant features, then passes these features through two, jointly trained recurrent units. The first recurrent cell uses the feature vector combined with the ground truth steering angles from the previous batch as input. The second recurrent cell uses the feature vector combined with the predicted steering angles from the previous batch as input. The weighted sum of the loss from both cells is used to update the network. During testing, only the recurrent cell with trained with the previous predicted angles is used.

We take a more simplified approach to Feudal Steering, as shown in Figure \ref{fig:fullNetwork}. Our network uses a 3D convolutional layer with a ReLU activation function followed by a dropout layer. The output of this convolution is saved to use later on in the network. This process is repeated four times before the output is fed through a series of fully connected layers with ReLU activation functions. At this point, the output and the intermediary representations from each of the convolutions are added together, passed through an ELU (exponential linear unit) layer, and normalized. Then, the previous predicted steering angle and the output of the ELU layer are passed through an LSTM. Finally, the output of the LSTM is passed through a fully connected layer with the output from the ELU layer to produce the steering angle. 

Compared to the Udacity network \cite{komanda}, we also use a set of 3D convolutional layers with ReLU, dropout regularization, and skip connections to glean relevant features from the images. However, we only train one LSTM with the concatenated feature vectors and previously predicted steering angle as input. Using the previous ground truth steering angle is feasible in the problem domain with the addition of extra sensors to the vehicle. However, our goal is to create a self-contained network that predicts steering angles based solely on image input, so we choose to use the previous predicted angle as input instead. Additionally, for a fully trained model, the difference between the previous ground truth and predicted angles will be negligible, so our performance at test time will not be greatly effected.

\begin{figure}[]
    \centering
    \includegraphics[width=.9\linewidth]{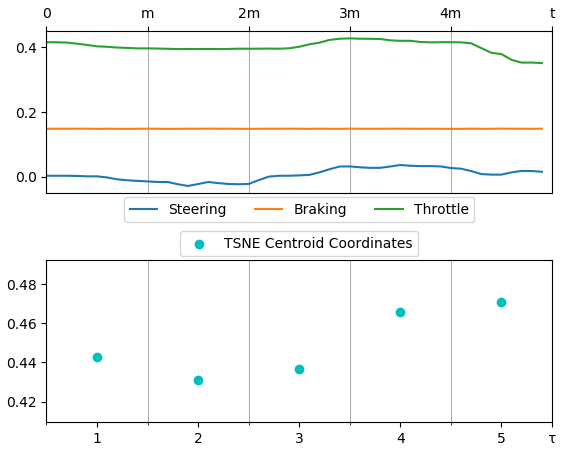}
    \caption{Steering, braking, and throttle data are concatenated every m time steps to make a vector of length 3m. Each vector 
    is projected to 2D t-SNE coordinates 
    that act like a manager for the steering angle prediction and operate at a lower temporal scale. In our experiments m=10; $t$ and $\tau$ are the temporal axes for the driving data and t-SNE coordinates respectively.}
    \label{fig:tsneBreakdown}

    \centering
    \includegraphics[width=.87\linewidth]{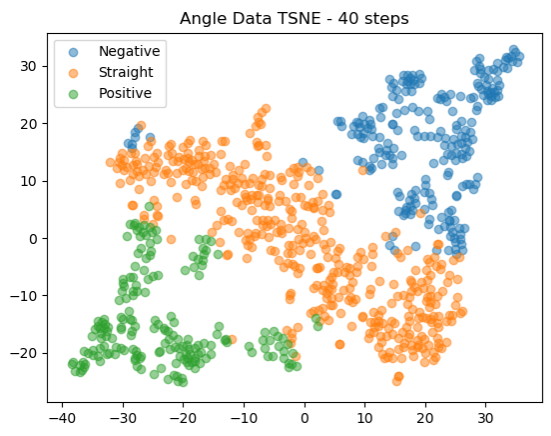}
    \caption{Total plot of the t-SNE coordinates for the Udacity data. The colors correspond to the average sign of the angles in each length 3m vector used to generate the points. The horizontal and vertical axes correspond to the two t-SNE dimensions.}
    \label{fig:tsnePlot}

    \centering
    \includegraphics[width=.87\linewidth]{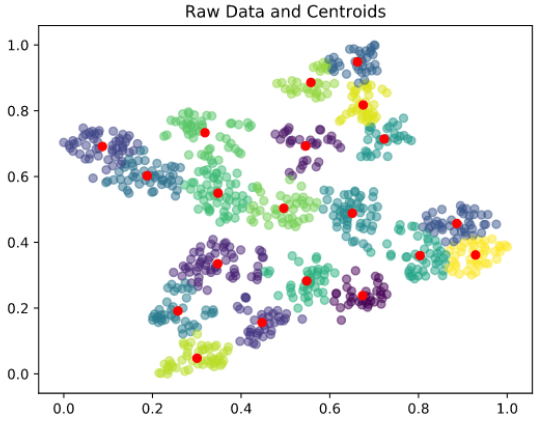}
    \caption{K-Means clustering (k=20) of the TSNE coordinates of the Udacity data with the centroids pictured in red. Not only do distinct clusters form in the data, but each cluster corresponds to a unique action of the vehicle.}
    \label{fig:kmeansCentroids}
\end{figure}

\begin{figure*}[]
    \centering
    \includegraphics[width=1.01\linewidth]{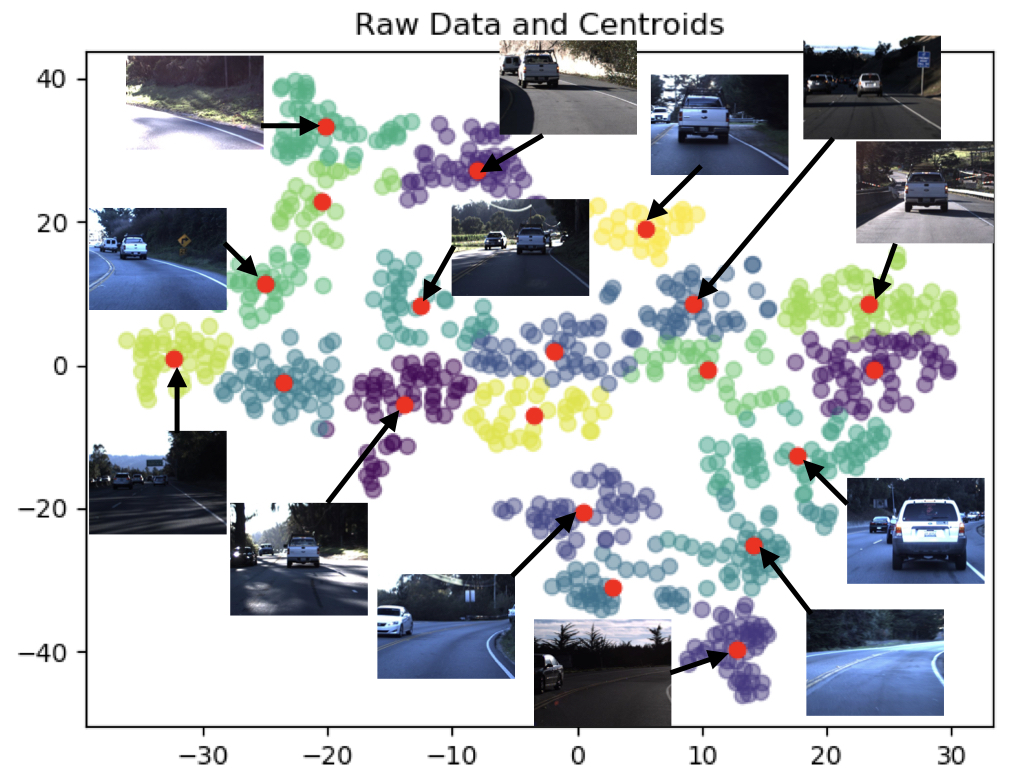}
    \caption{Example training images are shown with their corresponding t-SNE centroids. Notice the bottom right of the figure contains sharp right turns. Moving upwards and to the left, the right turn gets less sharp until the vehicle begins to go straight. Eventually this straight behavior starts to become a left turn until the vehicle is making sharp left turns in the upper left hand corner.}
    \label{fig:tsnePhotobomb}
\end{figure*}

\begin{figure}[]
    \centering
    \includegraphics[width=1.02\linewidth]{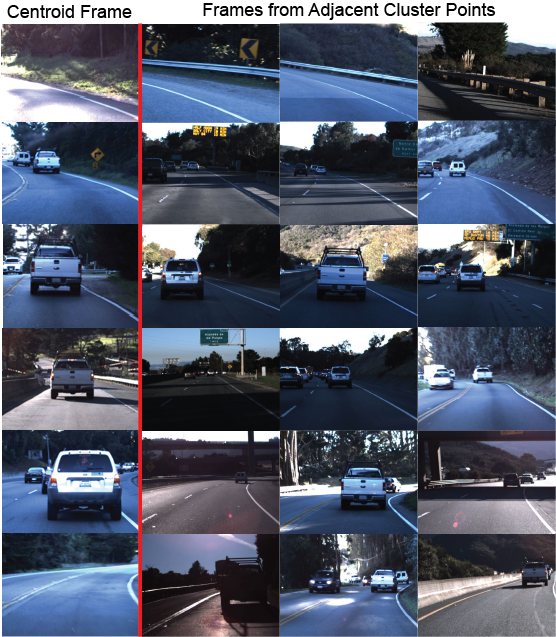}
    \caption{The left column of images are a subset of centroid frames from Figure \ref{fig:tsnePhotobomb}. The images to the right of each centroid frame come from different, adjacent points in the corresponding cluster for each centroid. Notice that the points in each cluster display similar behavior as their respective centroids. }
    \label{fig:clusterVer}
\end{figure}


\subsection{Subroutine ID}
For a hierarchical framework, we aim to classify the steering angles into their temporally abstracted subroutines, also called options or macro-actions, associated with highway driving
such as ``follow the sharp right bend'', ``bumper-to-bumper traffic'', ``bear left slightly''. 
This could be done by hand, but it would be a lengthy process, and the created subroutines would most likely be too simplistic to describe the wide variety of driving scenarios a vehicle may encounter. 
For driving, the high level tasks are numerous and it is preferable to compute or learn  {\it subroutine ids} rather than manually label semantic tasks. We demonstrate that our automatically extracted subroutine ids have observable semantic meaning in terms of driving tasks (see Figure~\ref{fig:tsnePhotobomb}).

\subsection{t-SNE Embedding as Subroutine ID}
We explore using t-SNE \cite{maaten2008visualizing} as an embedding space for our driving data and as the subroutine ids themselves. To do this, we arranged the steering angle, braking, and throttle pressure data into vectors of length $m$. Then, the vectors from each category that correspond to the same time steps are concatenated together to make vectors of length $3m$. During training, the collection of these vectors is passed through the unsupervised t-SNE algorithm to create a coordinate space for the driving data. For our networks, we use $m=10$, however this is a hyperparameter that can be tuned. 

Each vector of length $3m$ is given one x and y coordinate pair as illustrated in Figure \ref{fig:tsneBreakdown}. The greater collection of all of the generated points is shown in Figure \ref{fig:tsnePlot}. The coloring of the points in this figure is hard coded. The points corresponding to vectors with primarily negative steering angles are in blue. The points corresponding to vectors with positive steering angles are in green. The orange points correspond to vectors with steering angles that are relatively close to zero. 

Once we have the t-SNE embedding of the data, we use K-Means clustering on the coordinates and take the centroids of the clusters as our new subroutine ids, as shown in Figure \ref{fig:kmeansCentroids}. We vary k from ten to twenty to determine if different numbers of clusters improve prediction performance. Then, we train our manager network to predict subroutines similar to the t-SNE centroids given a sequence of images as input. In order to ensure that no data pertaining to the predicted steering angle is used as input to this network, we use the t-SNE centroid corresponding to the previous $m$ steering, braking, and throttle data as input to the network. 

To illustrate, refer back to Figure \ref{fig:tsneBreakdown}. If we are predicting an angle from the range $t\in[2m,3m]$, then the t-SNE centroid used for the subroutine id input to the angle prediction network will be the centroid at $\tau = 2$, which was made with the steering, braking, and throttle data from $t\in[m,2m]$. In this way, the angle we are attempting to predict will not be used to compute the t-SNE centroid that is input to the network as the subroutine id. This shift also incorporates an extra level of temporal abstraction into our network.

Figure \ref{fig:tsnePhotobomb} 
shows example training images that correspond to some of the t-SNE centroids. Notice that the bottom right of the figure contains sharp right turns. Moving diagonally upwards, the right turns get less sharp until the vehicle begins to go straight. Then, this straight motion gradually begins to become a left turn until, by the top left of the figure, the vehicle is making sharp left turns. Figure \ref{fig:clusterVer} shows that the points contained in each cluster exhibit the same, or comparable, behavior. The left column of images are a subset of the t-SNE centroid frames from Figure \ref{fig:tsnePhotobomb}. Each row contains frames from points adjacent to the associated centroid that are contained within the same cluster. The behavior in each row is consistent, showing that the points in each cluster behave similarly. 

\subsection{t-SNE Prediction Network}

Since our results (Section 4) show that t-SNE coordinates prove useful as a subroutine ID, we also explore prediction of t-SNE coordinates directly from images, as a t-SNE network following a concept introduced in prior work
\cite{xue2018deep}.  The t-SNE prediction network is jointly trained with and our steering angle prediction network. For this t-SNE manager network, we fine tune the FBResNet152 model \cite{he2016deep,cadene2018pretrained}. We train the steering angle prediction network to take in the predicted centroids as the subroutine id, as well as a sequence of images, in order to predict the next steering angle.

\subsection{Subroutine ID Prediction Network}

While t-SNE provides convenient visualization of the subroutine id semantic meaning, 
we take inspiration from \cite{kumar2019learning} to allow the manager learn the subroutines over the driving data.
This work trains multiple networks on completely unlabeled data in order to label frames based on an agent's actions during an initial exploration of an environment. The subroutines across these labeled frames are then learned and represented as discrete random variables. 
However, the Udacity dataset \cite{udacity} already provides low-level action labels between consecutive frames in the form of steering angles. So we only need to create a network to learn subroutines across these actions. 

In summary, we obtain subroutine ids using three methods:
1)  Set the subroutine id to the ground truth t-SNE cluster centroids where t-SNE is computed on steering, throttle, and braking data $m$ time steps prior to the prediction time $n$. 2) Set the subroutine id to the t-SNE network output following the general concept introduced in \cite{xue2018deep} by predicting t-SNE coordinates from images.  3) Learn subroutine ids jointly with steering angle prediction with a subroutine id network.
The best results are obtained by the third method.

\section{Experiments and Results}
\subsection{Dataset and Augmentation}
We test our feudal networks in the domain of autonomous vehicles using the Udacity driving dataset\cite{udacity}, which provides steering angles, first-person dash cam images, braking, and throttle pressure data. We use frames from the CH2\_002 partition of the dataset and use a 75\%/25\% train/test split. We augment our training data to increase its size and influence model training by implementing a horizontal flip, which effectively doubles the size of the dataset. For this change, we negate the angles associated with the flipped images. 
Additionally, all images are scaled and normalized so that their pixel values lie in the range $[-1,1]$.

\begin{figure}[]
    \centering
    \subfigure{\includegraphics[width=1.11\linewidth]{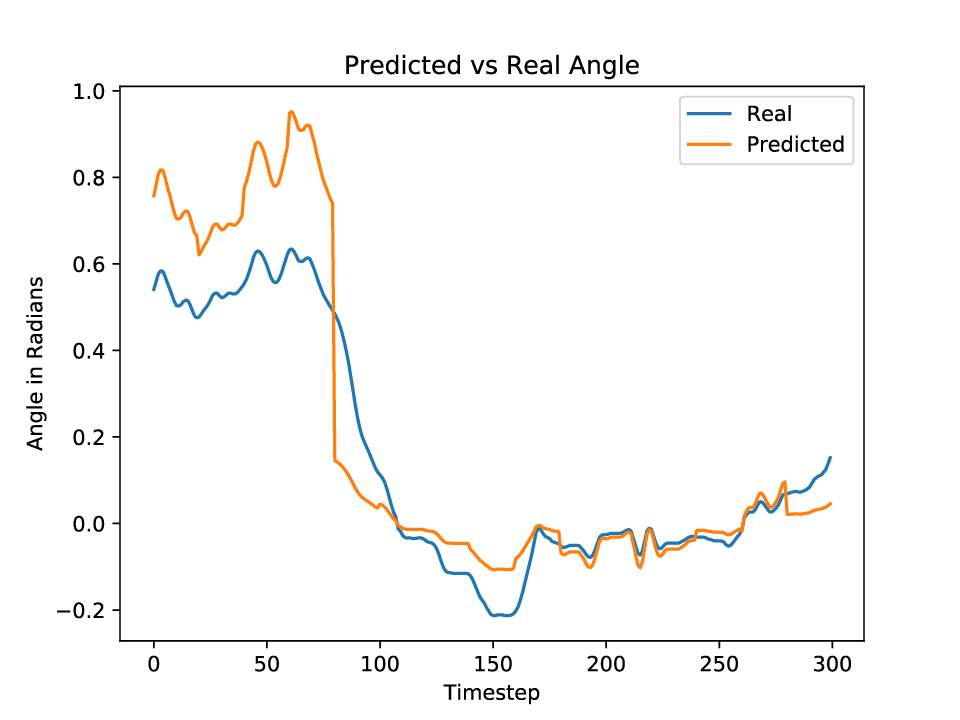}}
    
    \caption{Results of steering angle prediction when the ground truth t-SNE coordinates of the input data are used as the subroutine ids in the steering angle prediction network of Feudal Steering. Notice that, for these results, we use a network that does not take the previous angle as input.}
    \label{fig:tsneSubIDAngles}
\end{figure}

\subsection{t-SNE as Subroutine ID}
First, we use t-SNE as the embedding space for our subroutine ids by embedding the data into 2D space, using K-means clustering to create centroids, and using the coordinate pairs of those clusters as the subroutine ids. However, before we attempt to predict the t-SNE coordinates from the image data, we determine if the t-SNE coordinates will function as subroutine ids. We use the ground truth value of the t-SNE centroids as the subroutine id in our angle prediction network, along with an image sequence of length ten, to determine whether or not it would be worthwhile to attempt to predict the centroids. 

The results of this experiment are in Figure \ref{fig:tsneSubIDAngles}. The blue lines are the real steering angle, and the orange lines are the predicted angle. While the results in this figure show that the predicted angles diverge slightly from the ground truth angles, these predictions are more relevant to real world applications because they are computed using only visual input. Additionally, the quality of these predictions is high enough to motivate us to use additional methods of predicting the subroutine id's  with the manager network.

\subsection{Predicted t-SNE as Subroutine ID}
\begin{figure}[]
    \centering
    \includegraphics[width=1.11\linewidth]{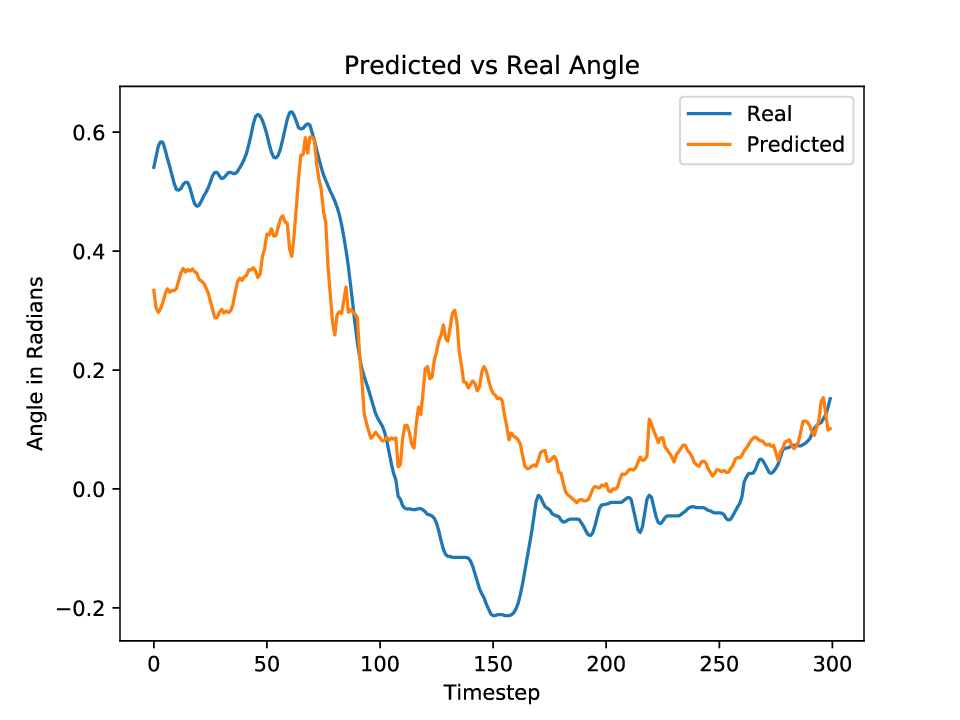}
    \caption{The steering angle prediction results using the predicted t-SNE network as the manager are pictured above in orange. The blue represents the ground truth angles. While these results as worse than our subroutine id netowrk, they were achieved without using the previous steering angle as input to the network.}
    \label{fig:predTSNEAng}
\end{figure}

Next, we jointly train a t-SNE prediction and steering angle prediction networks. The input to both is an image sequence of length ten. The t-SNE prediction network outputs the coordinates to the corresponding t-SNE centroid of the image input. To train this network, we minimize the MSE loss between the output and the ground truth t-SNE coordinates. The steering angle prediction network  takes in this predicted centroid and produces the corresponding steering angle. We also minimize the MSE loss between the predicted and real angles. We conducted this experiment using 10, 15, and 20 t-SNE centroids and found that 10 centroids produced the best results, as shown in Table \ref{tab:centroidComp}. Figure \ref{fig:predTSNEAng} shows the prediction results. The blue line represents the ground truth angles, and the orange line is the predicted angles. 

\begin{table}[]
    \centering
    \begin{tabular}{|c|c|c|c|}
        \hline
         Number of Centroids & 10 & 15 & 20 \\
         \hline 
         RMSE & 0.2093 & 0.2240 & 0.2251\\
         \hline 
    \end{tabular}
    \smallskip
    \caption{The manager network predicts subroutine ids to be close to the t-SNE centroids of the training data embedding space. We test 10, 15, and 20 centroids in our network and find that 10 centroids has the lowest RMSE.}
    \label{tab:centroidComp}
\end{table}

\subsection{Subroutine ID Network}

To create this subroutine id network, we mimic the structure of the steering angle network. However, the input to the subroutine id network is a one dimensional sequence of steering angles, so the network uses 1D convolutions instead of 3D. Additionally, we only use three sets of convolutions for this network instead of four. 

We jointly train the subroutine id and steering angle prediction networks. The subroutine id network takes a sequence of historical steering angles as input and outputs a goal vector representing the subroutine id for those angles. The steering angle network takes in the subroutine id, a sequence of images, and the previous predicted angle and outputs the next steering angle in the sequence. 

During training and testing, the sequence of angles fed into the subroutine id network consists of 
$[a_{n-1-m}, a_{n-1-(m-1)}, ..., a_{n-1}]$, in order to ensure that we only use the sequence of angles preceding the angle we aim to predict. The subroutine id is a single number that is able to take on any value in ${\rm I\!R}$. The sequence of images input to the steering angle network range from $[i_{n-m}, i_{n-(m-1)}, ..., i_n]$, and the previous angle used as input is $a_{n-1}$. We choose $m = 10$ for our experiments, but this is a hyperparameter that can be fine tuned.   

We use a learning rate of $1 \times 10^{-4}$ with an Adam optimizer. The other hyperparameters for the optimizer are unchanged from their pytorch defaults of $\beta = (.9, .999)$. We train our model under multiple loss functions and compare the performance. These loss functions are MSE,
\[\mathcal{L}_{MSE} = \frac{1}{N}\sum_{i=0}^N{( \alpha_i-a_i)^2}\]
RMSE, 
\[\mathcal{L}_{RMSE} = \sqrt{\frac{1}{N}\sum_{i=0}^N{( \alpha_i-a_i)^2}}\]
and MAE
\[\mathcal{L}_{MAE} = \frac{1}{N}\sum_{i=0}^N{| \alpha_i-a_i|}\]
where N is the number of predictions, $ \alpha$ is the ground truth angle, and $a$ is the predicted angle. We find that we achieve the best results using MSE loss, but we report our MAE loss for comparison purposes in the results section as well.

\begin{figure}[]
    \centering
    \subfigure{\includegraphics[width=1.1\linewidth]{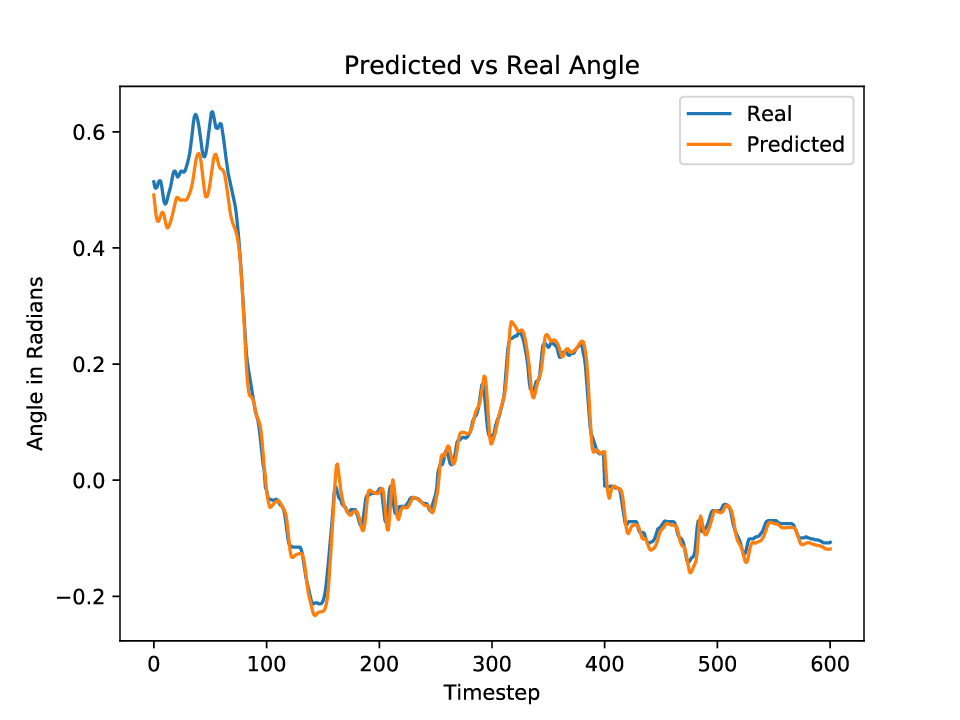}}
    
    \subfigure{\includegraphics[width=1.1\linewidth]{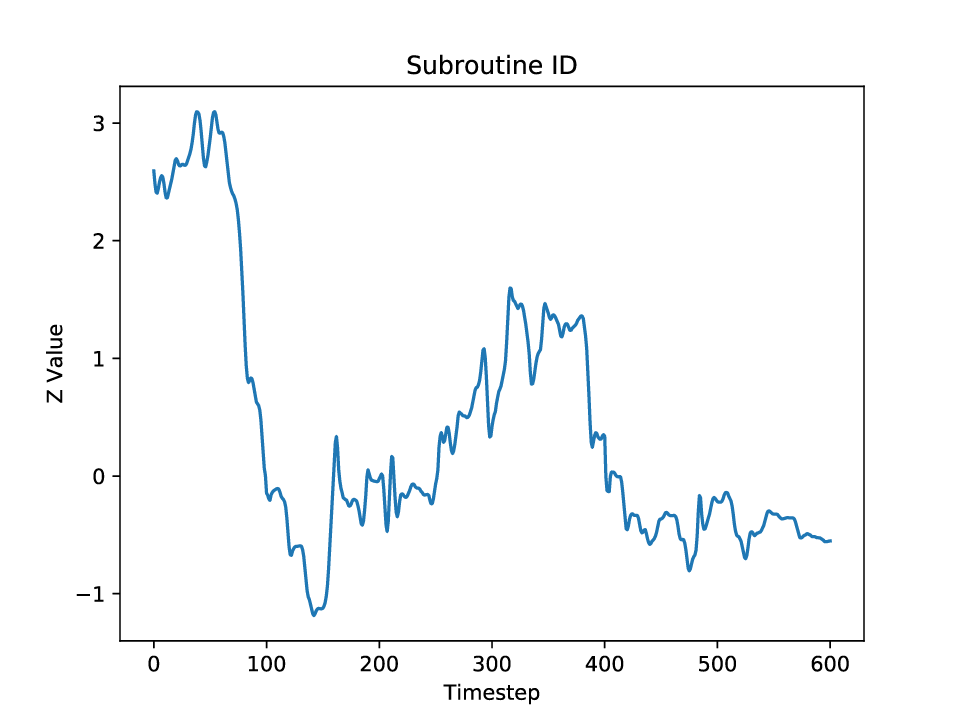}}
    \caption{Angle (top) and subroutine id (bottom) prediction results on the Udacity dataset obtained using our Feudal Steering network are shown above. The real angle is pictured in blue and the predicted angle is in orange. The subroutine ids are plotted alone. Notice that the subroutine id's behavior mimics the real angle behavior, but differs in scale.}
    \label{fig:predZK}
\end{figure}

Our final experiment is predicting steering angles and subroutines based on visual input using this subroutine id network. We create an image sequence of ten frames that we feed into our feudal network along with the previous steering angle to predict the next steering angle. Figure \ref{fig:predZK} shows the prediction results. The top graph shows the steering angle predictions. The corresponding subset of real steering angles from the Udacity \cite{udacity} dataset are in blue, and the predicted steering angles are in orange. 
The bottom graph in Figure \ref{fig:predZK} shows the predicted subroutine ids. 
We can see from these predictions that the learned subroutine ids follow the general pattern of the steering angles, but vary in scale, showing that the subroutine id is a stepping stone to the final steering angle prediction. 


We compare this method with several state of the art (SOTA) implementations in Table \ref{tab:rmseComp}. We show that our RMSE and MAE are lower than \cite{kim2017interpretable,maqueda2018event,chi2017deep,komanda}. While we did not achieve better loss values than \cite{hou2019learning}, we achieved comparable MSE and MAE values using a much smaller, simpler network. This is beneficial in the autonomous driving domain where memory and latency are limited for efficient, real world applications. 

\begin{table}[]
    \centering
    \begin{tabular}{|c|c|c|}
        \hline
          & RMSE & MAE\\
        \hline
        \hline
        Interpretable Attention \cite{kim2017interpretable} & - & 0.07191 \\
        \hline
        Event Based Camera\cite{maqueda2018event} & 0.07156 & - \\
        \hline
        Deep Steering \cite{chi2017deep} & 0.0609 & -\\
        \hline
        Udacity Challenge\cite{komanda}  &  0.0483 & -\\
        \hline
        \color{red}Feudal Steering (Ours)  &  \color{red}0.04659 & \color{red}0.01902 \\
        \hline
        Learning by Mimicking \cite{hou2019learning} & 0.04110 & 0.02834 \\ 
        \hline
    \end{tabular}
    \smallskip
    \caption{The RMSE and MAE of Feudal Steering is compared with other steering angle prediction methods. We outperform all but one of the SOTA methods. We produce comparable RMSE and MAE to \cite{hou2019learning} despite using a much smaller model. }
    \label{tab:rmseComp}
\end{table}

\subsection{Non-Hierarchical Steering Angle Prediction}
We attempted to use the steering angle prediction network without a manager network to compare hierarchical and non-hierarchical networks. However, the non-hierarchical network (worker network only) failed to predict any reasonably accurate steering angles. 

\section{Discussion}

In this work, we show that the feudal networks from  hierarchical reinforcement learning are more effective than reinforcement learning at the task of steering angle prediction. This effect is due to temporal abstraction. Breaking down the problem into more tractable pieces narrows the focus of the worker agent and allows the optimal policy to be found more quickly. Additionally, temporal abstraction also helps alleviate the problems of long term credit assignment and sparse reward signals. The lower temporal resolution of the manager shortens the period of time between rewards overall. 


We also explore 
a t-SNE embedding space 
as the goal space for the manager in our steering angle predictions. We use the centroid corresponding to steering angle, braking, and throttle data from the previous $m$ time steps as the subroutine id in our angle prediction network and were able to predict future steering angles without the direct use of the steering angle from the previous time step.  
However, this network had worse performance than our subroutine id network because of the limitations on the subroutine representation. When we allow the manager network the freedom to be able to define its subroutines for itself, performance increases and surpasses the current SOTA. 


\section*{Acknowledgements}
We acknowledge Lockheed Martin for support during this project. We thank Sanipa Arnold, Jeff Cammerata, and Matthew Purri for their suggestions and comments.
\clearpage
{\small
\bibliographystyle{ieee_fullname}
\bibliography{egbib}

\begin{thebibliography}{10}\itemsep=-1pt

\bibitem{bacon2017option}
Pierre-Luc Bacon, Jean Harb, and Doina Precup.
\newblock The option-critic architecture.
\newblock In {\em Thirty-First AAAI Conference on Artificial Intelligence},
  2017.

\bibitem{bojarski2016end}
Mariusz Bojarski, Davide Del~Testa, Daniel Dworakowski, Bernhard Firner, Beat
  Flepp, Prasoon Goyal, Lawrence~D Jackel, Mathew Monfort, Urs Muller, Jiakai
  Zhang, et~al.
\newblock End to end learning for self-driving cars.
\newblock {\em arXiv preprint arXiv:1604.07316}, 2016.

\bibitem{cadene2018pretrained}
Remi Cadene.
\newblock Pretrained models for pytorch.
\newblock {\em URL https://github. com/Cadene/pretrained-models. pytorch},
  2018.

\bibitem{chen2018deep}
Jianyu Chen, Zining Wang, and Masayoshi Tomizuka.
\newblock Deep hierarchical reinforcement learning for autonomous driving with
  distinct behaviors.
\newblock In {\em 2018 IEEE Intelligent Vehicles Symposium (IV)}, pages
  1239--1244. IEEE, 2018.

\bibitem{chi2017deep}
Lu Chi and Yadong Mu.
\newblock Deep steering: Learning end-to-end driving model from spatial and
  temporal visual cues.
\newblock {\em arXiv preprint arXiv:1708.03798}, 2017.

\bibitem{dayan1993feudal}
Peter Dayan and Geoffrey~E Hinton.
\newblock Feudal reinforcement learning.
\newblock In {\em Advances in neural information processing systems}, pages
  271--278, 1993.

\bibitem{duan2020hierarchical}
Jingliang Duan, Shengbo Li, Yang Guan, Qi Sun, and Cheng Bo.
\newblock Hierarchical reinforcement learning for self-driving decision-making
  without reliance on labeled driving data.
\newblock {\em IET Intelligent Transport Systems}, 2020.

\bibitem{komanda}
Ilya Edrenkin.
\newblock Komanda team solution, udacity challenge 1st place winner.
\newblock
  https://github.com/udacity/self-driving-car/blob/master/steering-models/community-models/komanda/solution-komanda.ipynb,
  2016.

\bibitem{haarnoja2018latent}
Tuomas Haarnoja, Kristian Hartikainen, Pieter Abbeel, and Sergey Levine.
\newblock Latent space policies for hierarchical reinforcement learning.
\newblock {\em arXiv preprint arXiv:1804.02808}, 2018.

\bibitem{he2016deep}
Kaiming He, Xiangyu Zhang, Shaoqing Ren, and Jian Sun.
\newblock Deep residual learning for image recognition.
\newblock In {\em Proceedings of the IEEE conference on computer vision and
  pattern recognition}, pages 770--778, 2016.

\bibitem{he2018aggregated}
Sen He, Dmitry Kangin, Yang Mi, and Nicolas Pugeault.
\newblock Aggregated sparse attention for steering angle prediction.
\newblock In {\em 2018 24th International Conference on Pattern Recognition
  (ICPR)}, pages 2398--2403. IEEE, 2018.

\bibitem{hou2019learning}
Yuenan Hou, Zheng Ma, Chunxiao Liu, and Chen~Change Loy.
\newblock Learning to steer by mimicking features from heterogeneous auxiliary
  networks.
\newblock In {\em Proceedings of the AAAI Conference on Artificial
  Intelligence}, volume~33, pages 8433--8440, 2019.

\bibitem{khan2019latent}
Qadeer Khan, Torsten Sch{\"o}n, and Patrick Wenzel.
\newblock Latent space reinforcement learning for steering angle prediction.
\newblock {\em arXiv preprint arXiv:1902.03765}, 2019.

\bibitem{kim2017interpretable}
Jinkyu Kim and John Canny.
\newblock Interpretable learning for self-driving cars by visualizing causal
  attention.
\newblock In {\em Proceedings of the IEEE international conference on computer
  vision}, pages 2942--2950, 2017.

\bibitem{kulkarni2016hierarchical}
Tejas~D Kulkarni, Karthik Narasimhan, Ardavan Saeedi, and Josh Tenenbaum.
\newblock Hierarchical deep reinforcement learning: Integrating temporal
  abstraction and intrinsic motivation.
\newblock In {\em Advances in neural information processing systems}, pages
  3675--3683, 2016.

\bibitem{kumar2019learning}
Ashish Kumar, Saurabh Gupta, and Jitendra Malik.
\newblock Learning navigation subroutines by watching videos.
\newblock {\em arXiv preprint arXiv:1905.12612}, 2019.

\bibitem{liang2018cirl}
Xiaodan Liang, Tairui Wang, Luona Yang, and Eric Xing.
\newblock Cirl: Controllable imitative reinforcement learning for vision-based
  self-driving.
\newblock In {\em Proceedings of the European Conference on Computer Vision
  (ECCV)}, pages 584--599, 2018.

\bibitem{lin2018architectural}
Shih-Chieh Lin, Yunqi Zhang, Chang-Hong Hsu, Matt Skach, Md~E Haque, Lingjia
  Tang, and Jason Mars.
\newblock The architectural implications of autonomous driving: Constraints and
  acceleration.
\newblock In {\em Proceedings of the Twenty-Third International Conference on
  Architectural Support for Programming Languages and Operating Systems}, pages
  751--766, 2018.

\bibitem{liu2019end}
Shikun Liu, Edward Johns, and Andrew~J Davison.
\newblock End-to-end multi-task learning with attention.
\newblock In {\em Proceedings of the IEEE Conference on Computer Vision and
  Pattern Recognition}, pages 1871--1880, 2019.

\bibitem{maaten2008visualizing}
Laurens van~der Maaten and Geoffrey Hinton.
\newblock Visualizing data using t-sne.
\newblock {\em Journal of machine learning research}, 9(Nov):2579--2605, 2008.

\bibitem{maqueda2018event}
Ana~I Maqueda, Antonio Loquercio, Guillermo Gallego, Narciso Garc{\'\i}a, and
  Davide Scaramuzza.
\newblock Event-based vision meets deep learning on steering prediction for
  self-driving cars.
\newblock In {\em Proceedings of the IEEE Conference on Computer Vision and
  Pattern Recognition}, pages 5419--5427, 2018.

\bibitem{mnih2013playing}
Volodymyr Mnih, Koray Kavukcuoglu, David Silver, Alex Graves, Ioannis
  Antonoglou, Daan Wierstra, and Martin Riedmiller.
\newblock Playing atari with deep reinforcement learning.
\newblock {\em arXiv preprint arXiv:1312.5602}, 2013.

\bibitem{nachum2018data}
Ofir Nachum, Shixiang~Shane Gu, Honglak Lee, and Sergey Levine.
\newblock Data-efficient hierarchical reinforcement learning.
\newblock In {\em Advances in Neural Information Processing Systems}, pages
  3303--3313, 2018.

\bibitem{rafati2019learning}
Jacob Rafati and David~C Noelle.
\newblock Learning representations in model-free hierarchical reinforcement
  learning.
\newblock In {\em Proceedings of the AAAI Conference on Artificial
  Intelligence}, volume~33, pages 10009--10010, 2019.

\bibitem{silver2017mastering}
David Silver, Thomas Hubert, Julian Schrittwieser, Ioannis Antonoglou, Matthew
  Lai, Arthur Guez, Marc Lanctot, Laurent Sifre, Dharshan Kumaran, Thore
  Graepel, et~al.
\newblock Mastering chess and shogi by self-play with a general reinforcement
  learning algorithm.
\newblock {\em arXiv preprint arXiv:1712.01815}, 2017.

\bibitem{song2019diversity}
Yuhang Song, Jianyi Wang, Thomas Lukasiewicz, Zhenghua Xu, and Mai Xu.
\newblock Diversity-driven extensible hierarchical reinforcement learning.
\newblock In {\em Proceedings of the AAAI Conference on Artificial
  Intelligence}, volume~33, pages 4992--4999, 2019.

\bibitem{sutton1999between}
Richard~S Sutton, Doina Precup, and Satinder Singh.
\newblock Between mdps and semi-mdps: A framework for temporal abstraction in
  reinforcement learning.
\newblock {\em Artificial intelligence}, 112(1-2):181--211, 1999.

\bibitem{tessler2017deep}
Chen Tessler, Shahar Givony, Tom Zahavy, Daniel~J Mankowitz, and Shie Mannor.
\newblock A deep hierarchical approach to lifelong learning in minecraft.
\newblock In {\em Thirty-First AAAI Conference on Artificial Intelligence},
  2017.

\bibitem{udacity}
Udacity.
\newblock Udacity self-driving car driving data 10/3/2016
  (dataset-2-2.bag.tar.gz).

\bibitem{vezhnevets2017feudal}
Alexander~Sasha Vezhnevets, Simon Osindero, Tom Schaul, Nicolas Heess, Max
  Jaderberg, David Silver, and Koray Kavukcuoglu.
\newblock Feudal networks for hierarchical reinforcement learning.
\newblock In {\em Proceedings of the 34th International Conference on Machine
  Learning-Volume 70}, pages 3540--3549. JMLR. org, 2017.

\bibitem{xu2017end}
Huazhe Xu, Yang Gao, Fisher Yu, and Trevor Darrell.
\newblock End-to-end learning of driving models from large-scale video
  datasets.
\newblock In {\em Proceedings of the IEEE conference on computer vision and
  pattern recognition}, pages 2174--2182, 2017.

\bibitem{xue2018deep}
Jia Xue, Hang Zhang, and Kristin Dana.
\newblock Deep texture manifold for ground terrain recognition.
\newblock In {\em Proceedings of the IEEE Conference on Computer Vision and
  Pattern Recognition}, pages 558--567, 2018.

\end{thebibliography}
}

\end{document}